\documentclass[11pt,a4paper]{article}
\usepackage[hyperref]{acl2020}
\usepackage{times}
\usepackage{latexsym}

\usepackage{graphicx}
\usepackage[ruled, linesnumbered]{algorithm2e}
\usepackage{amssymb,mathtools}
\usepackage{xcolor}
\usepackage{comment}
\usepackage{mathtools}
\usepackage[thinlines]{easytable}
\usepackage{babel,blindtext}

\usepackage{diagbox}
\usepackage{multirow}
\usepackage{longtable}
\usepackage{dblfloatfix}
\usepackage{scrextend}
\usepackage{float}
\usepackage{subcaption}
\usepackage{url}
\usepackage{amsfonts}

\usepackage{array}
\newcolumntype{L}[1]{>{\raggedright\let\newline\\\arraybackslash\hspace{0pt}}m{#1}}
\newcolumntype{C}[1]{>{\centering\let\newline\\\arraybackslash\hspace{0pt}}m{#1}}
\newcolumntype{R}[1]{>{\raggedleft\let\newline\\\arraybackslash\hspace{0pt}}m{#1}}
\usepackage{microtype}

\aclfinalcopy 

\title{TechTexC: Classification of Technical Texts using Convolution and Bidirectional Long Short Term Memory Network}
\author{Omar Sharif{\textdagger}, Eftekhar Hossain{*}, and Mohammed Moshiul Hoque{\textdagger}\\
 {\textdagger}Department of Computer Science and Engineering \\
  {*}Department of Electronics and Telecommunication Engineering\\
  Chittagong University of Engineering and Technology, Bangladesh \\
  \texttt{\textsuperscript{{\textdagger}}\{omar.sharif, moshiul\_240\}@cuet.ac.bd}\\
   \texttt{{\textsuperscript{*}}eftekhar.hossain@cuet.ac.bd}\
  }
  
\date{}

\begin{document}
\maketitle
\begin{abstract}
This paper illustrates the details description of technical text classification system and its results that developed as a part of participation in the shared task TechDofication 2020. The shared task consists of two sub-tasks: (i) first task identify the coarse-grained technical domain of given text in a specified language and (ii) the second task classify a text of computer science domain into fine-grained sub-domains. A classification system (called 'TechTexC') is developed to perform the classification task using three techniques: convolution neural network (CNN), bidirectional long short term memory (BiLSTM) network, and combined CNN with BiLSTM. Results show that CNN with BiLSTM model outperforms the other techniques concerning task-1 of sub-tasks (a, b, c and g) and task-2a. This combined model obtained $f_1$ scores of 82.63 (sub-task a), 81.95 (sub-task b), 82.39 (sub-task c), 84.37 (sub-task g), and 67.44 (task-2a) on the development dataset. Moreover, in the case of test set, the combined CNN with BiLSTM approach achieved that higher accuracy for the subtasks 1a (70.76\%), 1b (79.97\%), 1c (65.45\%), 1g (49.23\%) and 2a (70.14\%). 
\end{abstract}

\section{Introduction}
Due to the substantial growth and effortless access to the Internet in recent years, an enormous amount of unstructured textual contents have generated. It is a crucial task to organize or structure such a voluminous unstructured text in manually. Thus, automatic classification can be useful to manipulate a huge amount of texts, and extract meaningful insights which save a lot of time and money. Text categorization is a classical NLP problem which aims to categorize texts into organized groups. It has a wide range of applications like machine translation, question answering, summarization, and sentiment analysis. There are several approaches available to classify texts according to their labels. However, deep learning method outperforms the rule-based and machine learning-based models because of their ability to capture sequential and semantic information from texts \cite{minaee2020deep}. We propose a classifier using CNN \citep{jacovi2018understanding}, and BiLSTM \citep{zhou2016text} to classify technical texts in the computer science domain. Furthermore, by sequentially adding these networks, remarkable accuracy in several shared classification tasks can be obtained. The rest of the paper is organized as follows: related work given in section 2. Section 3 describes the dataset. The framework described in section 4. The findings presented in section 5. 

\section{Related Work}
CNN and LSTM have achieved great success in various NLP tasks such as sentence classification, document categorization, sentiment analysis, and summarization. \citet{kim2014convolutional} used convolution neural network to classify sentences. A method used contents and citations to classify scientific document \citep{cao2005doc}. \citet{zhou2016text} used 2-D max pooling and bidirectional LSTM to classify texts. \citet{zhou2015c} combined CNN and LSTM to classify sentiment and question type. Their system achieved superior accuracy than CNN and LSTM individually. \citet{hossain2020sentilstm} used LSTM to classify sentiment of Bengali text documents. Their system got maximum accuracy with one layer of LSTM followed by three dense layers. \citet{ranjan2017document} proposed a document classification framework using LSTM and feature selection algorithms. \citet{ameur2020robust} combined CNN and RNN methods to categorize Arabic texts. They used dynamic, fine-tuned words embedding to get effective result on open-source Arabic dataset.  

\section{Dataset}
To develop the classifier model, we used the dataset provided by the organizers of the shared task\footnote{https://ssmt.iiit.ac.in/techdofication.html}. This shared task consists of two subtasks: subtask-1 and subtask-2. Subtask-1 aims to the identification of coarse-grained domain for a piece of text. Organizers provided data including eight different languages (English, Bangla, Hindi, Gujarati, Malayalam, Marathi, Tamil, Telugu) each having a different number of classes for this task. In subtask-2, the goal is to find the fine-grained sub domain of a text from the computer science domains. Seven classes such as artificial intelligence, algorithm, computer architecture, computer networks, database management systems, programming, software engineering are available in this subtask-2. The number of training, validation and test texts for each of the task is different. Summary of the dataset presents in table \ref{data-table}.
\begin{table}[h!]
\centering
\begin{tabular}{C{1.2cm} C{1cm} C{1.2cm}C{1cm}C{1cm}}
\hline
\textbf{Task} & \textbf{No. of classes} & \textbf{Train} & \textbf{Dev}& \textbf{Test} \\
\hline
task-1a & 5 & 23962 & 4850 & 2500\\
task-1b & 5 & 58500 & 5842 & 1923\\
task-1c & 5 & 36009 & 5724 & 2682\\
task-1d & 7 &  148445 & 14338 & 4211\\
task-1e & 3 & 40669 & 3390 & 1514\\
task-1f & 4 & 41997 & 3780 & 1788\\
task-1g & 6 & 72483 & 6190 & 2070\\
task-1h & 6 & 68865 & 5920 & 2611\\
task-2a & 7 & 13580 & 1360 & 1929\\
\hline
\end{tabular}
\caption{Dataset description}
\label{data-table}
\end{table}


\section{System Overview}
\begin{figure*}[ht!]
\centering
  \includegraphics[height=4.5cm, width=14cm]{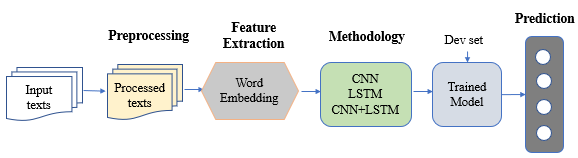}
  \caption{Schematic diagram of our system}
  \label{fig:proposed_model}
\end{figure*}

Figure~\ref{fig:proposed_model} shows the schematic diagram of the proposed system. 
The system has four major parts: preprocessing, feature extraction, classifier model and prediction. After processing the raw texts, Word2Vec word embedding technique is applied on the processed texts to extract features. After exploiting inherent features of the texts, the model trained with CNN, BiLSTM and combination of CNN \& BiLSTM.. Finally, the trained model will use to predict the class on the development set. 

\subsection{Preprocessing}
In this step, all the punctuation's (,.;:"!) and flawed characters (\#,\$, \%,*,@) removed from the input texts. Texts are having a length of fewer than two words also discarded. Deep learning algorithms could not possibly learn from the raw texts. Thus, a numeric mapping of the input texts is created. A vocabulary of $K$ unique words is developed and each input text encoded into numeric sequences based on word index in vocabulary. By applying the pad sequence method, each sequence converted into fixed-length vector. We choose optimal sequence length 100 as most of the length of the text ranges between 30-70 words. In order to maintain a fixed length of inputs, zero paddings are used with the short text, and extra values discarded from the long sequences.

\subsection{Feature Extraction}
To extract features from texts and capture semantic property of a word Word2Vec \cite{mikolov2013distributed} embedding technique is used. Embedding maps textual data into a dense vector by solving the sparsity problem. We use the default embedding layer of Keras to produce embedding matrix. Embedding layer has three parameters: vocabulary size, embedding dimension and length of texts. Embedding dimension determine the size of the dense word vector. The entire corpus is fitted into the embedding layer for a specific subtask and choose 100 as embedding dimension for all the subtasks. Features extracted from the embedding layer propagated the rest of the network. 

\subsection{Classifier Model}
In this work, CNN  and BiLSTM are used for initial model building. However, after combining these methods, we get superior results in several subtasks \citep{zhou2015c}. A description of the proposed architecture illustrates in the subsequent paragraphs.

\paragraph{CNN:} In CNN, convolution filters capture the inherent syntactic and semantic features of the texts. The proposed classifier considers two layers, one dimensional CNN. In each layer, there are 128 filters with kernel size 5. To downsample the features on CNN max-pooling technique is utilized where pool size is $1\times 5$. We have used a non-linear activation function `relu' with CNN.

\paragraph{BiLSTM:} We use Bidirectional LSTM network to capture the sequential features from the input text and to avoid vanishing/exploding gradient problems of simple RNN. We use two layers of BiLSTM on top of each other, where each layer has 128 LSTM cells. In order to reduce the overfitting on training data, the dropout technique is used with a dropout rate of 0.2. After achieving the hidden representation form, the LSTM layer output passed to the softmax layer for classification.

\paragraph{CNN+BiLSTM:} In this approach, we merge CNN and BiLSTM models with marginal modification in network architecture. Previously, we used two layers of CNN and BiLSTM, whereas in this technique, discard one layer from each network and combine them sequentially. Word embedding features is feed to the CNN, which has 128 filters. After max pooling with a window of size 5, features of CNN propagated to the LSTM layer. It has 128 bidirectional cells to capture the sequential information. In order to mitigate overfitting, a dropout layer is added with a dropout rate of 0.2. Finally, the softmax layer gets input from the LSTM and perform classification.   

\subsection{Prediction}
The goal of the prediction module is to determine the technical domain of an input text that it has never seen before. For the prediction, sample instances are processed and converted into numerical sequences by the tokenizer. Trained model use this sequence to predict the associated class of the input text.

\section{Experiments}
Google co-laboratory platform is used to conduct experiments. Deep learning model developed with Keras=2.4.0 framework with tensorflow=2.3.0 in the backend. For data preparation and evaluation, we use python=3.6.9 and secikit-learn=0.22.2.

\subsection{Hyperparameter Settings}
Performance of deep learning models heavily depends on the hyperparameters used in training. To choose the optimal hyperparameters for the proposed model, we played with different combinations. We choose parameter values based on its effect on the output. Table \ref{hyperparameter} exhibits the values of different hyperparameters considered to train the proposed model. \begin{table}[h!]
\centering
\begin{tabular}{lc}
\hline
\textbf{Hyperparameters} & \textbf{Optimum value}\\
\hline
Embedding dimension & 100\\
Padding length & 100\\
Filters & 128\\
Kernel size & 5\\
Pooling type & max \\
Window size & 5\\
LSTM cell & 128\\
Dropout rate & 0.2\\
Optimizer & `adam'\\
Learning rate & 0.001\\
Batch size & 128\\
\hline
\end{tabular}
\caption{\label{hyperparameter}
Hyperparameter Settings
}
\end{table}
\noindent
Adam optimizer is used with a learning rate of $0.001$. The model trained with a batch size of 128 until a training accuracy of 98\% reached. We use Keras callbacks to save the intermediate model during training with best validation accuracy. The trained model used to predict on the instances of development set.

\subsection{Results}
We determine the superiority of the models based on their weighted $f_1$ score on the development set of different tasks. Table \ref{result} shows the evaluation results of CNN, BiLSTM and CNN+BiLSTM. 
\begin{table*}[h!]
\centering
\begin{tabular}{l|ccc|ccc|ccc}
\hline
\textbf{Task} & \multicolumn{3}{c}{\textbf{CNN}}& \multicolumn{3}{c}{\textbf{Bi-LSTM}}& \multicolumn{3}{c}{\textbf{CNN+Bi-LSTM}}\\
\hline
&\textbf{P}&\textbf{R}&\textbf{F}&\textbf{P}&\textbf{R}&\textbf{F}&\textbf{P}&\textbf{R}&\textbf{F}\\
\hline
task-1a (English) &81.48 & 81.36 & 81.4 &82.81 & 82.52 & 82.52 & 82.9 & 82.54 & \textbf{82.63} \\
task-1b (Bangla) &81.96 &81.94 &81.91 &81.49 &81.38 &81.39 &82.04 &81.97 & \textbf{81.95} \\
task-1c ( Gujarati) &82.63 &82.39 &82.38 &82.79 &82.05 &82.06 &82.58 &82.41 & \textbf{82.39} \\
task-1d (Hindi) &79.46 &79.0 &79.07 &79.86 &79.54 &\textbf{79.52} &79.65 &79.44 &79.48 \\
task-1e (Malayalam) &91.51 &91.53 &91.52 &91.87 &91.89 &\textbf{91.86} &91.38 &91.36 &91.32 \\
task-1f (Marathi) &85.93 &85.93 &85.84 &86.47 &86.53 &\textbf{86.48} &86.57 &86.51 &86.38 \\
task-1g (Tamil) &84.26 &84.07 &84.02 &84.36 &84.34 &84.3 &84.56 &84.63 &\textbf{84.37} \\
task-1h (Telegu) &86.85 &86.82 &86.78 &87.64 &87.34 &\textbf{87.41} &87.17 &87.14 &87.13\\
task-2a (English) &64.36 &63.82 &63.83 &66.45 &65.51 &65.72 &67.86 &67.35 &\textbf{67.44} \\
\hline
\end{tabular}
\caption{\label{result} Evaluation results of three models on different tasks where P, R, F denotes precision, recall and weighted $f_1$ score.
}
\end{table*}
\begin{table*}[h!]
\centering
\begin{tabular}{l|c|cccc}
\hline
\textbf{Task} & \textbf{Method}& \textbf{A}
&\textbf{P}&\textbf{R}&\textbf{F}\\
\hline
task-1a (English)& CNN+BiLSTM &70.76 & 71.50 &	70.76 &	70.63 \\
task-1b (Bangla)&CNN+BiLSTM &79.97&	81.50 &82.41 &80.25\\
task-1c ( Gujarati) &CNN+BiLSTM&65.45 &1.95 &	1.81 &1.86 \\
task-1d (Hindi) &BiLSTM&57.28 & 57.13 & 55.99 &	54.57\\
task-1e (Malayalam)&BiLSTM &31.37 &0.32 &0.18 &0.19 \\
task-1f (Marathi) &BiLSTM&63.09&65.98& 61.38 &59.81\\
task-1g (Tamil) &CNN+BiLSTM&49.23&48.38 &61.34 &43.70\\
task-1h (Telegu)& BiLSTM &52.82&0.76& 0.64&0.68\\
task-2a (English) &CNN+BiLSTM&70.14 & 71.51 &	70.19 &	70.40\\
\hline
\end{tabular}
\caption{\label{result_test} Evaluation results on the test set. Here A, P, R, F denotes accuracy, precision, recall and weighted $f_1$ score respectively.}
\end{table*}
The results revealed that BiLSTM model achieved the higher $f_1$ score of $79.52\%, 91.86\%, 86.48\%$ and $87.41\%$ for tasks 1d, 1e, 1f and 1h. It outperforms CNN model for all tasks. The reason behind the superior results of LSTM because of its capability to capture long-range dependencies. However, combined CNN and BiLSTM provide interesting insights. It outdoes previous BiLSTM model in tasks 1a, 1b, 1c, 1g and 2a by obtaining $82.63\%, 81.95\%, 82.39\%, 84.37\%$ and $67.44\%$ $f_1$ scores. The model achieved 2\% rise in $f_1$ score concerning task-2a where the fine-grained domain of a text is identified. In all the cases, there exists a small difference $(<0.5\%)$ between the result of BiLSTM and CNN+BiLSTM. By analyzing the results, it observed that for a task with less number of classes, all models achieved quite similar performance. However, when the number of classes increased, the BiLSTM and CNN+BiLSTM models performed better than CNN. It is because the CNN model could not capture sequential feature as well compare to LSTM.

Table \ref{result_test} shows the output of the best run on the test set for each tasks. Based on the performance of the development set, methods are selected to predict on the test set. Therefore, we use CNN+BiLSTM model to predict on the tasks 1a, 1b, 1c, 1g and 2a. Model achieved $70.63\%, 80.25\%, 1.86\%, 43.70\%$ and $70.4\%$ weighted $f_1$ scores on these tasks respectively. Unlike other tasks, precision, recall, and $f_1$ score are much lower for task 1c compare to the validation results. This lower score might happen due to some mistake during evaluation. Task 2a get better $f_1$ score on the test set to compare to the development set. For other cases, the performance of the methods degraded on the development set.

BiLSTM method used to get the outputs for tasks 1d, 1e, 1f and 1h. Model obtained 57.13\%, 0.32\%, 65.98\% and 0.76\% weighted $f_1$ scores on these tasks. It suspected that some errors might occur during evaluation for tasks 1e and 1h. The model achieved 91.86\% and 87.41\% $f_1$ scores on these tasks for the validation set but got an implausible result on the test set. This error might occur due to Unicode issues of different languages. Our system also encountered an error when data read from the text file. The performance of BiLSTM method decreased in the test set than the validation set for all tasks.  

Precision, recall and $f_1$ score have fallen for each task in the test set except task 2a. Weighted $f_1$ score has increased by 2.5\% in the test set. For all the tasks, we observed a substantial variation between the development set and test set results. There might be two possible reasons behind this unpredictable nature of the models. First one, model is overfitted on the training set. Thus, it gets better results on training and validation set but poor results on the test set. The second one, test data are more diverse than training data. Suppose significant overlap does not exist between the train and test features. In that case, the model indeed performs poor on the test data since the models learn from the characteristics of training data.

\section{Conclusion}
This paper presents a detail description of the proposed system and its evaluation for the technical texts classification in different languages. As the baseline method, we used CNN and BiLSTM, and compare these methods with the proposed model (combined CNN and BiLSTM). Each model is trained, tuned and evaluated separately for subtasks 1 and 2. The proposed method showed better performance in terms of accuracy for subtasks (a, b, c, g) of task 1 and task 2a on development set. However, in the case of test set, the system performed better for the subtasks 1a, 1b, 1c, 1g and 2a. More dataset can be included for improved performance. In future, the attention mechanism may be explored to observe its effects on text classification tasks.

\bibliography{anthology,acl2020}
\bibliographystyle{acl_natbib}

\end{document}